\documentclass[conference]{IEEEtran}
\IEEEoverridecommandlockouts

\usepackage[T1]{fontenc}
\usepackage{cite}
\usepackage{amsmath,amssymb,amsfonts}
\usepackage{algorithm}
\usepackage{algorithmic}
\usepackage{graphicx}
\usepackage{textcomp}
\usepackage[table]{xcolor}
\usepackage{booktabs}
\usepackage{multirow}
\usepackage{array}
\usepackage{pifont}
\usepackage{subcaption}
\usepackage{url}
\usepackage{hyperref}

\definecolor{BestPurple}{RGB}{203,187,222}
\definecolor{SecondPurple}{RGB}{235,229,244}

\newcommand{\best}[1]{\cellcolor{BestPurple}#1}
\newcommand{\second}[1]{\cellcolor{SecondPurple}#1}

\newcommand{\besttext}[1]{\colorbox{BestPurple}{\strut #1}}
\newcommand{\secondtext}[1]{\colorbox{SecondPurple}{\strut #1}}

\newtheorem{remark}{Remark}

\begin{document}

\title{One-Shot Hierarchical Federated Clustering for Distributed Non-IID Biomedical Data Analysis}

\author{
\IEEEauthorblockN{Shenghong Cai}
\IEEEauthorblockA{
\textit{Guangdong Provincial Key Laboratory IRADS}\\ 
\textit{and Department of Computer Science}\\
\textit{Beijing Normal-Hong Kong Baptist University}, Zhuhai, China\\
w630201601@mail.bnbu.edu.cn
}
\and
\IEEEauthorblockN{Yiqun Zhang\textsuperscript{*}}
\IEEEauthorblockA{
\textit{School of Future Technology}\\
\textit{South China University of Technology}, Guangzhou, China\\
yqzhangzyq@gmail.com\\
\textsuperscript{*}Corresponding author
}
}

\maketitle

\begin{abstract}
In multi-institutional biomedical data analysis, comprehensive disease profiles are often fragmented across different data silos. Furthermore, local data distributions typically exhibit varying levels of granularity or even complex nested structures. Although Hierarchical Clustering (HC) is a powerful tool for exploring such data topologies, its cumbersome recursive process incurs prohibitive computational overhead and renders it highly susceptible to privacy leakage. This paper, therefore, introduces an efficient one-shot hierarchical FC framework that performs client-end (institutions) distribution exploration and server-end distribution aggregation through one-way prototype-level communication from clients to the server. Although the designed one-shot prototype transmission guarantees minimal computational overhead and mitigates privacy risks, it inevitably incurs information loss. To counteract this, a multi-granular competitive learning mechanism is proposed on the server to fuse the clusterlets, effectively maximizing the retention and mining of global distribution topologies under limited, one-way communication constraints. It turns out that the complex cluster distributions across clients can be efficiently explored, and extensive experiments comparing state-of-the-art methods on public datasets demonstrate the superiority of the proposed framework in balancing robust clustering, privacy preservation, and computational efficiency. 
Code and online appendix are available at \url{https://github.com/hongdream/FedHCA}.

\end{abstract}


\section{Introduction}
Multi-institutional bioinformatics analysis plays an important role in discovering disease patterns, identifying latent patient subgroups, and supporting collaborative biomedical knowledge discovery~\cite{brisimi2018federated,ye2023adaptive,fan2025clustering}. However, biomedical data are often collected and maintained by different institutions, where strict privacy requirements and institutional data-sharing constraints make data centralization impractical. Federated clustering (FC) provides a promising unsupervised paradigm for this purpose, aiming to obtain global cluster distributions by leveraging clients' data while preserving privacy~\cite{yin2021comprehensive,chung2022federated,zhang2025federated}. 
However, due to the non-IID nature of data distributed across clients~\cite{qin2023fedapen,chen2024lightweight} and the absence of label guidance, aligning divergent local cluster distributions to achieve a coherent global clustering objective remains challenging~\cite{zou2024federated,zhang2026one}. The presence of local clusters at varying granularities further exacerbates this issue, as misalignment across granular levels can severely degrade global clustering quality\cite{li2020federated,wei2020federated}.

\begin{figure}[!t]
    \centering
    \includegraphics[width=\linewidth]{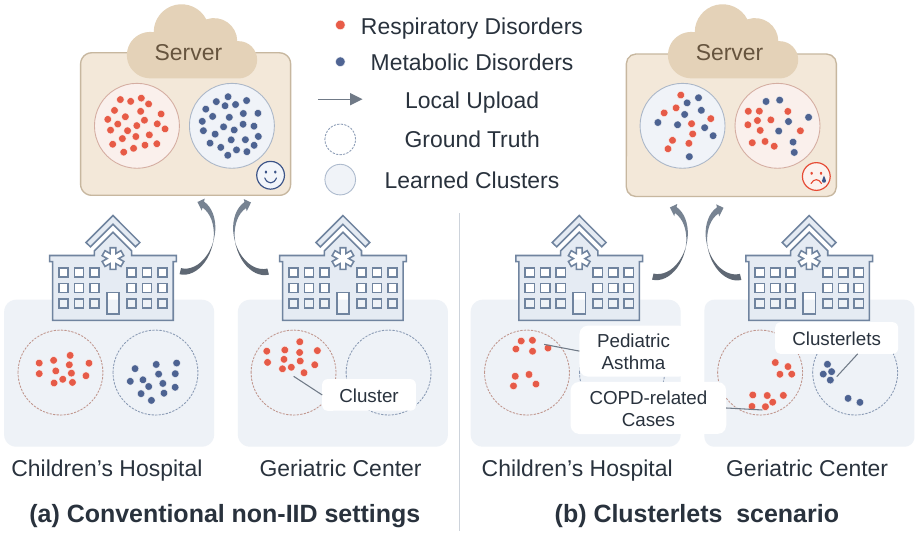}
    \caption{\textbf{Conventional non-IID settings vs. Clusterlets scenario.} (a) relies on a typical non-IID assumption that the complete distribution of clusters within each client can be sufficiently reflected. By contrast, (b) shows a more realistic scenario that clusters are fragmented into several clusterlets and distributed across different clients, which can easily lead to serious distortions at the server in aggregating global clusters.}
    \label{fig:intro_diagram}
\end{figure}

Most existing FC methods allow clients to learn local cluster distributions and share privacy-preserving knowledge with the server for global aggregation.
For instance, FedSC~\cite{qiao2024federated} and FFCM~\cite{stallmann2022towards} require multiple communication rounds to iteratively refine global clusters, incurring high overhead and privacy risks. 
By contrast, \cite{dennis2021heterogeneity} proposes a one-shot communication scheme that reduces bandwidth and latency while improving privacy.

% 这里修改了example，图1的画法应与这里的example联系
Notably, many existing FC methods are developed under a relatively conventional non-IID setting, where the data on each client are assumed to reflect the distribution of certain complete clusters sufficiently (see Fig.~\ref{fig:intro_diagram}(a)). 
However, in realistic federated scenarios, a global cluster may be fragmented across clients into incomplete local sub-clusters with different granularities, which we refer to as clusterlets (see Fig.~\ref{fig:intro_diagram}(b)). 
For example, in collaborative disease discovery, a broad respiratory-related disease category may be fragmented across hospitals, where each client observes only partial data objects of the global category. 
These local observations can differ substantially, such as pediatric asthma cases in a children's hospital and COPD-related cases in a geriatric center. 
Consequently, local clients may form distinct local clusters from their limited observations, making it difficult for global aggregation to recognize that these heterogeneous local clusters, i.e., clusterlets, may belong to the same global disease-related cluster.
Such incomplete and granularity-misaligned local observations make it difficult to aggregate local clustering results into a coherent global clustering structure.

A promising approach to capturing the inherent characteristics of data is to construct a hierarchy over data objects~\cite{cai2024robust}. However, due to privacy constraints in FC, conventional Hierarchical Clustering (HC) methods~\cite{murtagh2012algorithms} are unsuitable, as they typically operate at the object-level and cannot simultaneously support cluster-level aggregation and capture the unveiled multi-granular cluster structure. Therefore, the fragmentation of global clusters across clients presents two key challenges in the FC setting: 1) The presence of incomplete clusters further complicates the exploration of different client-specific distributions, hindering the aggregation process, and 2) The misalignment in cluster granularity levels across clients impedes hierarchical structure aggregation at the server side.

In this paper, we propose a Federated Hierarchical Clusterlet Aggregation (FedHCA) algorithm, adopting a one-shot communication framework.
First, each client applies Fine-grained Clusterlet Discovery (Fine-CD) to adaptively estimate the number of clusterlets, capturing the local clusterlet distributions. To preserve privacy, only the centroids of clusterlets are transmitted to the server via a one-shot, one-way communication scheme. Subsequently, the server applies Multi-granular Clusterlet Aggregation (Multi-CA) to integrate the received centroids and automatically uncover the inherent global cluster distributions across multiple granularities, thereby constructing a cluster-level hierarchical structure. Both Fine-CD and Multi-CA adopt a unified competitive penalized learning mechanism that drives candidate clusters to compete for data objects while gradually eliminating low-importance clusters, which explores the optimal cluster distributions without prior knowledge. To generate final partitional clustering results, the learned multi-granular structural information is embedded into data-enhanced representations, effectively bridging the granularity gap across distributed clients. Consequently, these modules collaboratively constitute a novel one-shot hierarchical federated clustering framework, offering a robust solution to reconcile local structural fragmentation and global distribution alignment in privacy-sensitive biomedical scenarios. The major contributions of this work are as follows:

\begin{itemize}
    %\item We propose a novel FC framework that captures distinct distributions and automatically aggregates multi-granular local clusters with the global clustering objective, enhancing robustness under structural inconsistencies. 
    \item We propose a novel one-shot FC framework that reconciles the tension between privacy preservation and structural information loss. By synergizing one-way prototype transmission with multi-granular competitive learning, it overcomes the privacy and computational bottlenecks, maximizing the extraction of global topologies.

    \item This paper addresses a more realistic yet challenging non-IID scenario in biomedical FC, where global clusters are fragmented into multiple clusterlets distributed across heterogeneous clients, yielding diverse local distributions.

    \item A hierarchical aggregation mechanism is developed to integrate local clusterlets across multiple granularity levels, thereby constructing a global hierarchical structure based on the underlying multi-granular cluster distributions.
\end{itemize}

\section{Related Work}
\subsection{Federated Clustering}
FC with multiple communication rounds includes methods like IFCA~\cite{ghosh2020efficient}, which enables unsupervised clustering by maintaining separate models per cluster~\cite{zhang2024prototype}. 
FedSC~\cite{qiao2024federated} approximates global similarity matrices via kernel decomposition to handle non-IID client data. 
Fed-FCM~\cite{pedrycz2021federated} iteratively updates cluster prototypes, while FFCM~\cite{stallmann2022towards} refines cluster centers based on local data to address fragmentation and heterogeneity. 
Recently, AFCL~\cite{zhang2025asynchronous} further adapts by asynchronously learning global structures without assuming the number of clusters.
However, these methods require multiple communication rounds for model aggregation or refinement, incurring high overhead and elevating privacy risks~\cite{wang2025one,mahadik2020fast}.

In contrast, one-shot methods~\cite{dennis2021heterogeneity} address above concerns by performing clustering in a single communication round, sharing only cluster-level statistics, e.g., local centroids, minimizing the risk of information leakage and adversarial attacks. 
Approaches such as $k$Fed~\cite{dennis2021heterogeneity} perform local clustering at clients and upload cluster centers for global aggregation. 
OSFSC~\cite{xie2023fed} uses a one-shot framework for subspace clustering by learning a unified affinity matrix from distributed data. 
NN-FC~\cite{wang2024one} further employs Laplacian noise to ensure privacy. 
However, they overlook the exploration of the multi-granular and nested data distribution.

\subsection{Hierarchical Clustering}
HC is a promising way to understand inherent characteristics of data~\cite{da2025treediffusion,zhang2023learning}, as it produces a hierarchical structure of data distribution by recursively linking the most similar object pairs.
Despite common linkage strategies~\cite{murtagh2012algorithms}, advanced strategies have been explored for scalability~\cite{dogan2022k,zhang2025adaptive}.
Beyond linkage-based HC, hierarchical Bayesian models have
also been developed for grouped data clustering~\cite{fan2022grouped,fan2021unsupervised}.
Nevertheless, traditional HC methods focus on addressing object-level hierarchical structure, requiring access to raw data, which hampers their applicability in privacy-sensitive scenarios under the FC settings~\cite{he2024co}, where raw data sharing is prohibited.

\begin{figure*}[!t]
    \centering
    \includegraphics[width=1\linewidth]{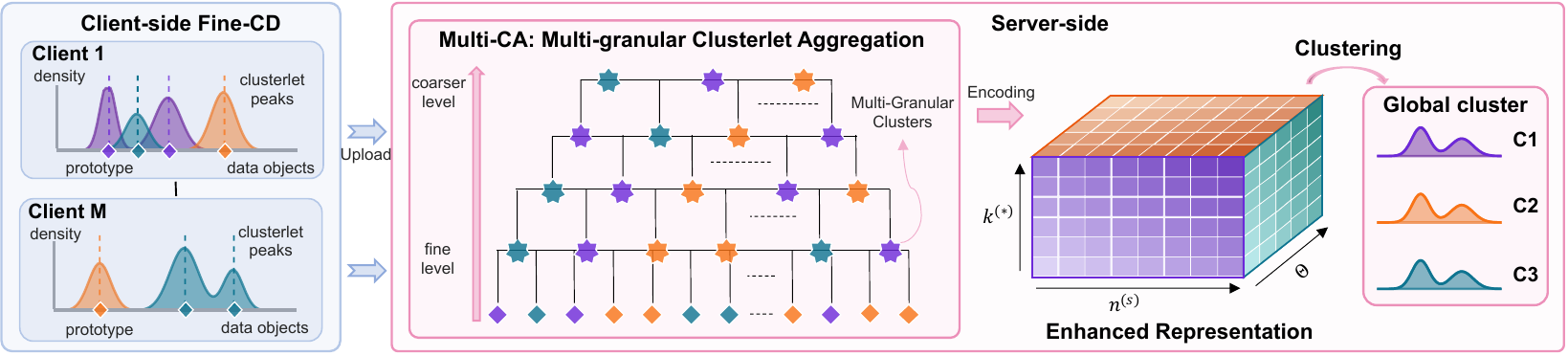}
    \caption{Overview of FedHCA. Each client performs Fine-CD to extract local clusterlets and uploads only their centroids to the server. The server then executes Multi-CA to derive a multi-granular hierarchy in a bottom-up manner. The resulting multi-granular distributions are embedded into enhanced representations and are used to conduct the final clustering.}
    \label{fig:method_diagram}
\end{figure*}

\section{Proposed Method}

We first formalize the FC problem and then introduce two core components of the proposed FedHCA: 1) clients perform Fine-CD to explore clusterlets in the presence of incomplete clusters, and 2) the server performs Multi-CA to uncover multi-granular cluster distributions and construct a hierarchical structure. FedHCA further embeds multi-granular cluster information into data-enhanced representations for informative clustering. The overview of FedHCA is shown in Fig.~\ref{fig:method_diagram}.

The FC problem considered in this paper is first defined as follows.
Considering a federated network with $M$ clients, the global dataset is distributed across these clients, where the $m$-th client holds a private dataset $\mathbf{X}^{(m)} \in \mathbb{R}^{n^{(m)} \times D}$ comprising $n^{(m)}$ data objects with $D$ features. These local datasets collectively constitute the global dataset, satisfying $\sum_{m=1}^{M} n^{(m)} = N$.

Given these distributed local datasets, the objective of FC is to collaboratively learn a global cluster structure while keeping raw data locally stored. Specifically, each client infers the affiliation of its data to $k^*$ global clusters $C = \{ C_j \}_{j=1}^{k^*}$ by maximizing the intra-cluster similarity:
\begin{equation}
\label{eq:objective_function}
P(\mathbf{Q}, C) = \sum_{i=1}^{N} \sum_{j=1}^{k^*} q_{ij} s(\mathbf{x}_i, C_j),
\end{equation}
where $\mathbf{Q}\in\mathbb{R}^{N\times k^*}$ is the object-cluster assignment matrix, whose entries satisfy $q_{ij} \in \{ 0, 1 \}$ and $\sum_{j=1}^{k^*} q_{ij} = 1$.
The detailed definition and computation of the object-cluster similarity $s(\mathbf{x}_i,C_j)$ are provided in the Online Appendix.

\subsection{Fine-CD: Fine-grained Clusterlet Discovery on the Clients}
\label{subsection:Fine-CD}
To discover the incomplete clusters as clusterlets on the client side, we propose a Fine-grained Clusterlet Discovery (Fine-CD) algorithm, which initializes a large number of candidate clusterlets $C^{(m)} = \{ C^{(m)}_j \mid 1 \le j \le k_0 \}$ with a substantially large $k_0$. 
These clusterlets are designed to compete with each other, ultimately capturing fine-grained data distributions by preserving prominent clusterlets while eliminating redundant ones. 

For the $m$-th client, we define a clusterlet weight set $W = \{ w_j \mid 1 \le j \le k_0 \}$, where $w_j$ denotes the weight of clusterlet $C^{(m)}_j$ in the clustering structure. 
For each object $\mathbf{x}^{(m)}_i$, we define the competitive score of clusterlet $C^{(m)}_j$ as
\begin{equation}
\label{eq:competitive_score}
\psi_{ij}^{(m)}
=
\lambda_j w_j s(\mathbf{x}^{(m)}_i, C^{(m)}_j).
\end{equation}
The winning clusterlet $C_v^{(m)}$ is selected as the clusterlet with the largest competitive score:
\begin{equation}
\label{eq:v}
v
=
\operatorname*{arg\,max}_{1 \le j \le k_0}
\psi_{ij}^{(m)}.
\end{equation}
The rival clusterlet $C_r^{(m)}$ is then selected as the most competitive clusterlet among the remaining candidates:
\begin{equation}
\label{eq:r}
r
=
\operatorname*{arg\,max}_{1 \le j \le k_0,\, j \neq v}
\psi_{ij}^{(m)}.
\end{equation}
In Eq.~(\ref{eq:competitive_score}), $\lambda_j$ is the relative winning possibility of $C_j^{(m)}$, computed as:
\begin{equation}
\label{eq:lambda}
\lambda_j=1-\frac{g_j}{ {\textstyle \sum_{t=1}^{k_0}g_t + 1} }.
\end{equation}
Here, $g_j$ records the winning times of $C^{(m)}_j$ in the learning process, updated by $g_v = g_v + 1$.

While $\lambda_j$ acts as a penalty mechanism to ensure fair competition, the clusterlet weight $w_j$ defined in Eq.~(\ref{eq:competitive_score}) must be dynamically updated to reflect the importance of each clusterlet. To constrain the values of clusterlet weight $w_j$ within the interval $[0, 1]$, we introduce an intermediate variable $\tau_j$ to update $w_j$. Specifically, we reward the winner $C^{(m)}_v$ by:
\begin{equation}
\label{eq:v_weight}
\tau_v = \tau_v + \eta,
\end{equation}
and penalize the nearest rival $C^{(m)}_r$ by decreasing $\tau_r$:
\begin{equation}
\label{eq:client_penalize}
\tau_r = \tau_r - \eta \frac{s(\mathbf{x}^{(m)}_i, C^{(m)}_r)}{s(\mathbf{x}^{(m)}_i, C^{(m)}_v)},
\end{equation}
where $\eta$ denotes the learning rate. Through this competitive mechanism, the winner $C^{(m)}_v$ is rewarded with a small step $\eta$, making it more likely to be preserved as a dominant clusterlet. Conversely, the rival $C^{(m)}_r$ is adaptively penalized by the similarity ratio, meaning that a rival highly similar to the winner will be heavily suppressed.
Then, a Sigmoid function smoothly regularizes $\tau_j$ into clusterlet weight $w_j$ by:
\begin{equation}
\label{eq:client_weight}
w_j = \frac{1}{1+e^{-10(\tau_j + 5)}}.
\end{equation}

It can be noted that once the clusterlet weight $w_j$ approaches 0, the corresponding clusterlet is eliminated and its associated data objects will be reassigned to the remaining clusterlets. In this way, the $m$-th client ultimately obtains $k^{(m)}$ clusterlets. To avoid directly sharing raw data, the $m$-th client uploads only the clusterlet centroids set $C^{(m)} = \{ \mathbf{c}^{(m)}_j \mid 1 \le j \le k^{(m)} \}$ to the server for global aggregation. 
\begin{remark}
\textbf{Prototype-level Communication:}
    FedHCA requires one round of communication, with each local client uploading clusterlet centroids $C^{(m)}$ to the server, ensuring that only a minimal amount of local information is shared across the federated network. Existing privacy-preserving techniques, such as homomorphic encryption \cite{acar2018survey} and differential privacy \cite{li2023differentially}, can be integrated into FedHCA to enhance data confidentiality during the transmission of $C^{(m)}$. 
\end{remark}

\subsection{Multi-CA: Multi-granular Clusterlet Aggregation}
\label{subsection:Multi-CA}
At this stage, the server receives clusterlet centroids from all clients, denoted as $C = \{ C^{(m)} \mid 1 \le m \le M \}$. By stacking all local prototype sets $C^{(m)}$, we obtain a global matrix $\mathbf{C} \in \mathbb{R}^{n^{(s)} \times D}$, where $n^{(s)}$ is the total number of clusterlet centroids.

The Multi-granular Clusterlet Aggregation (Multi-CA) algorithm is proposed to automatically aggregate multi-granular global distributions and build a hierarchy.
To establish the foundation of this hierarchy, Multi-CA initiates the clustering process at the finest level of granularity. Starting from a significantly large initial cluster number $k_0$, Multi-CA applies the competitive penalized learning process (Eqs.~(\ref{eq:competitive_score})-(\ref{eq:client_weight})) to partition $\mathbf{C}$ into $k_1$ clusters, yielding the finest-grained prototype-cluster affiliation matrix $\mathbf{Q}_1$.
To explore coarser-grained distributions, the process is recursively applied, each round initialized by inheriting the cluster number from the previously converged stage (e.g., $k_1$) and resetting all internal learning factors, i.e., relative winning possibility $\lambda_j \ (1 \le j \le k_1)$ and cluster weights $W$. The recursion terminates when no meaningful coarser structure can be identified.
In the end, the server systematically captures the global distribution from the finest to the coarsest granularity levels, yielding a sequence of monotonically decreasing cluster numbers (i.e., $k_1 > k_2 > \cdots > k_{\Theta}$).
By aggregating the outcomes from all stages, Multi-CA constructs a multi-granular hierarchy $\mathcal{H}$, where each hierarchical level corresponds to a prototype-cluster affiliation structure under a specific granularity:
\begin{equation}
\label{eq:hierarchy}
\mathcal{H}
=
\left\{
(\mathbf{Q}_{\theta}, k_\theta)
\right\}_{\theta=1}^{\Theta},
\quad
k_1 > k_2 > \cdots > k_\Theta,
\end{equation}
where $\mathbf{Q}_{\theta} \in \mathbb{R}^{n^{(s)} \times k_\theta}$ denotes the prototype-cluster affiliation matrix at the $\theta$-th hierarchical level, $k_\theta$ is the corresponding number of clusters, and $\Theta$ represents the total number of hierarchical levels.

\begin{remark}
\label{remark:inherit-k}
\textbf{Initialization Stability of Multi-CA:}
At each granularity level, Multi-CA reinitializes cluster centroids under the cluster number inherited from the previous stage. 
Although this design encourages the exploration of diverse multi-granular structures, random centroid initialization may introduce instability in the learned hierarchy. 
To further improve robustness, more stable initialization strategies, such as k-means++~\cite{arthur2007k} and scalable k-means++~\cite{bahmani2012scalable}, can be incorporated into Multi-CA.
\end{remark}

\subsection{Hierarchical Representation for Global Clustering}
\label{section:FedHCA}
To obtain a clustering result w.r.t. the target number of clusters $k^*$ based on the multi-granular hierarchy $\mathcal{H}$, FedHCA embeds the prototype-cluster affiliations from each layer $\mathbf{Q}_\theta \ (1 \le \theta \le \Theta)$ into a new data-enhanced representation $\mathbf{X}^{(s)} \in \mathbb{R}^{n^{(s)} \times \Theta}$. 
For an $n^{(s)} \times k_\theta$ affiliation matrix $\mathbf{Q}_\theta$, the $\theta$-th encoded feature of $\mathbf{x}^{(s)}_i$ is defined as the categorical value of its affiliated cluster at granularity level $\theta$:
\begin{equation}
\label{eq:encode}
x^{(s)}_{i\theta} = a,
\quad \text{if} \quad
q^{(\theta)}_{ia}=1,
\quad
a \in \{1,\ldots,k_\theta\}.
\end{equation}

Since the features in $\mathbf{X}^{(s)}$ are categorical, we employ a feature-weighted categorical similarity measure as follows:
\begin{equation}
\label{eq:similarity_server}
s(\mathbf{x}^{(s)}_i, C^{(s)}_j)
=
\sum_{\theta=1}^{\Theta}
\rho_{j\theta}
\mathbb{I}\left(x^{(s)}_{i\theta}=c^{(s)}_{j\theta}\right),
\end{equation}
where $\mathbb{I}(\cdot)$ returns $1$ if the condition holds and $0$ otherwise. 
To estimate $\rho_{j\theta}$ for categorical values, we consider both the inter-cluster separability $\xi_{j\theta}$~\cite{yuan2023spatio} and the intra-cluster compactness $\kappa_{j\theta}$. 
For $\theta$-th encoded feature, the empirical categorical distribution of cluster $C_j^{(s)}$ is constructed as:
\begin{equation}
\label{eq:categorical_distribution}
p_{j\theta,a}
=
\frac{1}{|C_j^{(s)}|}
\sum_{\mathbf{x}^{(s)}_i \in C_j^{(s)}}
\mathbb{I}\left(x^{(s)}_{i\theta}=a\right),
\end{equation}
where $a \in \{1,\ldots,k_\theta\}$ denotes a categorical value at the $\theta$-th granularity level. 
Similarly, the empirical categorical distribution of its complement $\bar{C}_j^{(s)}$ is denoted as $\bar{p}_{j\theta,a}$, which is constructed by replacing $C_j^{(s)}$ with $\bar{C}_j^{(s)}$ in Eq.~(\ref{eq:categorical_distribution}).

The inter-cluster separability $\xi_{j\theta}$ is computed as the Hellinger distance between the two empirical categorical distributions:
\begin{equation}
\label{eq:xi_server}
\xi_{j\theta}
=
\sqrt{
1-
\sum_{a=1}^{k_\theta}
\sqrt{
p_{j\theta,a} \cdot \bar{p}_{j\theta,a}
}
}.
\end{equation}
The intra-cluster compactness $\kappa_{j\theta}$ is measured by the concentration of the categorical distribution within $C_j^{(s)}$:
\begin{equation}
\label{eq:kappa_server}
\kappa_{j\theta}
=
\sum_{a=1}^{k_\theta}
p_{j\theta,a}^{2}.
\end{equation}
Finally, the feature-cluster weight $\rho_{j\theta}$ is calculated by:
\begin{equation}
\label{eq:server_feature_cluster_weight}
\rho_{j\theta} =
\frac{\xi_{j\theta}\kappa_{j\theta}}
{\sum_{t=1}^{\Theta}\xi_{jt}\kappa_{jt}}.
\end{equation}

To optimize the objective function in Eq.~(\ref{eq:objective_function}), FedHCA jointly optimizes $\mathbf{Q}^{(s)}$ and $\rho$ by alternately fixing one variable while updating the other.
After obtaining the converged $\mathbf{Q}^{(s)}$, each uploaded clusterlet prototype is assigned to one global cluster. The corresponding assignment is then mapped back to the local objects belonging to that clusterlet, thereby producing the final global clustering result for all distributed objects.

\section{Experiment}

\begin{table*}[!t]
    \centering
    \caption{Clustering performance. \besttext{Dark Purple} and \secondtext{Light Purple} denote the best and second-best results, respectively. ``$\overline{AR}(ST)$'' row reports the average performance ranks and the Wilcoxon signed-rank test results, where a significant difference between the corresponding method and FedHCA is indicated by ``+''.}
    \label{tbl:performance}
    %\resizebox{\textwidth}{!}{
    \begin{tabular}{c|c|c c c c c c c|c}
    \toprule
    \multirow{1}{*}{\textbf{Index}} & \multirow{1}{*}{\textbf{Data}} & \textbf{$k$Fed} & \textbf{FFCM-1} & \textbf{FFCM-2} & \textbf{OSFSC} & \textbf{FedSC} & \textbf{NN-FC} & \textbf{AFCL} & \textbf{FedHCA} \\
    \midrule
    \multirow{10}{*}{Purity}
    & EC & \second{0.687±0.01} & 0.417±0.05 & 0.374±0.02 & 0.538±0.03 & 0.345±0.00 & 0.650±0.06 & 0.348±0.01 & \best{0.780±0.05} \\
    & US & 0.470±0.01 & 0.465±0.00 & 0.459±0.01 & 0.409±0.02 & 0.343±0.00 & \second{0.473±0.04} & 0.337±0.00 & \best{0.529±0.05} \\
    & VE & \second{0.380±0.02} & 0.342±0.00 & 0.343±0.00 & 0.362±0.01 & 0.347±0.01 & 0.368±0.02 & 0.298±0.00 & \best{0.406±0.02} \\
    & EP & \second{0.345±0.01} & 0.341±0.00 & 0.331±0.00 & \best{0.348±0.01} & 0.342±0.00 & 0.334±0.00 & 0.341±0.00 & 0.324±0.00 \\
    & YE & \second{0.447±0.05} & 0.294±0.00 & 0.289±0.00 & 0.406±0.02 & 0.283±0.00 & 0.358±0.04 & 0.281±0.01 & \best{0.527±0.06} \\
    & CA & \second{0.515±0.03} & 0.375±0.00 & 0.351±0.00 & 0.467±0.02 & 0.441±0.05 & 0.418±0.02 & 0.289±0.00 & \best{0.600±0.04} \\
    & LA & 0.610±0.01 & 0.613±0.00 & 0.475±0.00 & 0.612±0.00 & 0.622±0.01 & \second{0.669±0.05} & 0.283±0.11 & \best{0.752±0.01} \\
    & WI & 0.492±0.02 & \best{0.520±0.00} & 0.464±0.00 & 0.485±0.01 & 0.467±0.01 & 0.467±0.00 & 0.464±0.00 & \second{0.514±0.02} \\
    & PE & 0.632±0.00 & 0.335±0.00 & 0.311±0.00 & \best{0.664±0.02} & 0.519±0.04 & 0.655±0.03 & 0.162±0.00 & \second{0.662±0.06} \\
    & LE & 0.335±0.01 & 0.143±0.00 & 0.100±0.00 & 0.329±0.01 & 0.241±0.02 & \second{0.338±0.02} & 0.089±0.00 & \best{0.346±0.02} \\
    \midrule
    \multicolumn{2}{c|}{$\overline{AR}(ST)$}
    & 2.9 (+) & 4.9 (+) & 6.6 (+) & 3.5 (+) & 5.3 (+) & 3.6 (+) & 7.5 (+) & 1.9 \\
    \midrule
    \multirow{10}{*}{ARI}
    & EC & \second{0.420±0.06} & 0.060±0.04 & 0.048±0.01 & 0.163±0.04 & -0.001±0.01 & 0.415±0.12 & 0.013±0.01 & \best{0.688±0.09} \\
    & US & \second{0.155±0.00} & 0.127±0.00 & 0.125±0.01 & 0.045±0.01 & 0.001±0.00 & 0.128±0.05 & 0.000±0.00 & \best{0.227±0.12} \\
    & VE & \second{0.079±0.00} & 0.048±0.00 & 0.049±0.00 & 0.037±0.00 & 0.061±0.00 & 0.056±0.02 & 0.000±0.00 & \best{0.088±0.02} \\
    & EP & 0.015±0.01 & 0.010±0.01 & \second{0.016±0.00} & 0.014±0.01 & 0.003±0.00 & 0.002±0.01 & 0.000±0.00 & \best{0.017±0.00} \\
    & YE & \second{0.198±0.06} & 0.045±0.00 & 0.044±0.00 & 0.088±0.01 & 0.003±0.00 & 0.097±0.04 & 0.003±0.01 & \best{0.245±0.06} \\
    & CA & \second{0.193±0.02} & 0.064±0.00 & 0.061±0.00 & 0.107±0.02 & 0.185±0.05 & 0.122±0.03 & 0.000±0.00 & \best{0.280±0.04} \\
    & LA & 0.383±0.02 & 0.372±0.00 & 0.219±0.00 & 0.315±0.01 & \second{0.483±0.03} & 0.452±0.05 & 0.041±0.13 & \best{0.539±0.05} \\
    & WI & 0.031±0.03 & \second{0.038±0.00} & -0.010±0.01 & 0.022±0.02 & 0.003±0.00 & -0.010±0.01 & 0.000±0.00 & \best{0.048±0.03} \\
    & PE & 0.457±0.00 & 0.137±0.00 & 0.153±0.01 & 0.426±0.04 & 0.368±0.04 & \second{0.511±0.03} & 0.000±0.00 & \best{0.516±0.07} \\
    & LE & \second{0.160±0.01} & 0.037±0.00 & 0.005±0.00 & 0.080±0.02 & 0.129±0.01 & 0.144±0.01 & 0.000±0.00 & \best{0.174±0.02} \\
    \midrule
    \multicolumn{2}{c|}{$\overline{AR}(ST)$}
    & 2.5 (+) & 5.1 (+) & 5.9 (+) & 4.9 (+) & 5.1 (+) & 4.0 (+) & 7.7 (+) & 1.0 \\
    \bottomrule
    \end{tabular}%}
\end{table*}

\begin{table}[!t]
    \centering
    \caption{Statistics of the datasets. The ``Category'' column indicates the broad data category of each dataset.}
    \label{tbl:dataset}
    \resizebox{\columnwidth}{!}{
    \begin{tabular}{c|c c c|c c c}
    \toprule
    \textbf{No.} & \textbf{Data} & \textbf{Abbrev.} & \textbf{Category} & \textbf{$n$} & \textbf{$d$} & \textbf{$k^*$}\\
    \midrule
    1  & Ecoli & EC & \textbf{Biomedical} & 336 & 7 & 8\\
    2  & User Knowledge Modeling & US & General & 403 & 5 & 4\\
    3  & Statlog(Vehicle Silhouettes) & VE & Recognition & 845 & 18 & 4\\
    4  & HCV for Egyptian Patients & EP & \textbf{Biomedical} & 1385 & 28 & 4\\
    5  & Yeast & YE & \textbf{Biomedical} & 1484 & 8 & 10\\
    6  & Cardiotocography & CA & \textbf{Biomedical} & 2126 & 21 & 10\\
    7  & Statlog(Landsat Satellite) & LA & Recognition & 6435 & 36 & 6\\
    8  & Wine Quality & WI & General & 6497 & 11 & 7\\
    9  & Pen-Based Digits & PE & Recognition & 10992 & 16 & 10\\
    10 & Letter Recognition & LE & Recognition & 20000 & 16 & 26\\
    11 & Synthetic Dataset 1 & SD1 & Synthetic & 100000 & 10 & 5\\
    12 & Synthetic Dataset 2 & SD2 & Synthetic & 500 & 10000 & 5\\
    \bottomrule
    \end{tabular}
    }
\end{table}

\subsection{Experimental Settings}
\textbf{Five Experiments} are conducted: 1) clustering performance evaluation, 2) clusterlets visualization, 3) ablation study, 4) parameter sensitivity analysis, and 5) scalability evaluation. Due to space limitations, the detailed scalability evaluation is provided in the 
\href{https://github.com/hongdream/FedHCA}{Online Appendix}.

\textbf{Seven SOTA Counterparts} are compared, including FedSC \cite{qiao2024federated}, FFCM-1 and FFCM-2 \cite{stallmann2022towards}, AFCL \cite{zhang2025asynchronous}, $k$Fed \cite{dennis2021heterogeneity}, OSFSC \cite{xie2023fed}, and NN-FC \cite{wang2024one}. The hyperparameters of these methods (if any) are set following their source papers. Based on the Parameter Sensitivity Analysis in section~\ref{sec:parameter}, we set the learning rate to $\eta = 0.05$ and the initial number of clusters is set to $k_0 = 0.5n^{(m)}$ for Fine-CD and $k_0 = 0.5n^{(s)}$ for Multi-CA. For all compared methods, the target number of clusters $k^*$ is set to the ground-truth number of classes for each dataset, as reported in Table~\ref{tbl:dataset}.

\textbf{Four Validity Indices} are utilized, including Purity, Adjusted Rand Index (ARI), Normalized Mutual Information (NMI), and Clustering Accuracy (ACC). Among them, Purity, NMI, and ACC are in the interval $[0, 1]$, while ARI is in the interval $[-1, 1]$. Higher values of these indices correspond to better clustering performance \cite{ezugwu2022comprehensive} .

\textbf{Twelve Datasets} are employed for a comprehensive evaluation, including 10 real-world datasets and two synthetic datasets. All real-world datasets are obtained from the UCI Machine Learning Repository. Among the 10 real-world datasets, four pertain to biomedical analysis, while the remaining six cover other application domains, examining FedHCA's generalizability beyond a single domain. The two synthetic datasets are further used to evaluate the scalability and efficiency of FedHCA under controlled settings. Detailed dataset statistics are shown in Table~\ref{tbl:dataset}. In the experimental simulations, we set the number of clients to $M = 8$.
% 详细说明分配的原则
The $k$-means algorithm is applied to each global cluster to generate several clusterlets. These clusterlets are subsequently randomly allocated to different clients to emulate local cluster distributions, reflecting the fragmented nature of real-world FC scenarios.

\subsection{Clustering Performance Evaluation}
The clustering performance of FedHCA and its counterparts is compared in Table~\ref{tbl:performance}, which reports results on the Purity and ARI metrics.
For each dataset, we run each method 10 times and report the mean and standard deviation.
The best and second-best results are marked in \besttext{Dark Purple} and \secondtext{Light Purple}, respectively.
The ``$\overline{AR}(ST)$'' row summarizes the average rank of each method across datasets, and the symbol ``(+)'' indicates a significant difference based on the two-sided Wilcoxon signed-rank test~\cite{benavoli2016should} at the 95\% confidence level.
\textbf{Observations}:
1) FedHCA outperforms the counterparts in most cases, highlighting its superiority. Although it ranks second in terms of Purity on WI and PE, the gaps are small, demonstrating its effectiveness across diverse scenarios.
2) On four biomedical datasets, i.e., EC, EP, YE, and CA, FedHCA achieves competitive performance in terms of Purity and ARI.
It obtains the best Purity and ARI on EC, YE, and CA.
On EP, although FedHCA achieves a lower Purity of 0.3244 compared with the best result of 0.348, it obtains the highest ARI of 0.0174.
These results indicate that FedHCA remains effective in biomedical-related scenarios, despite performance variations across datasets.
3) The significance tests confirm that FedHCA demonstrates statistically significant differences from all counterparts across the evaluated datasets.

\subsection{Clusterlets Visualization}
\begin{figure}[!t]
    \centering
    \includegraphics[width=0.49\linewidth]{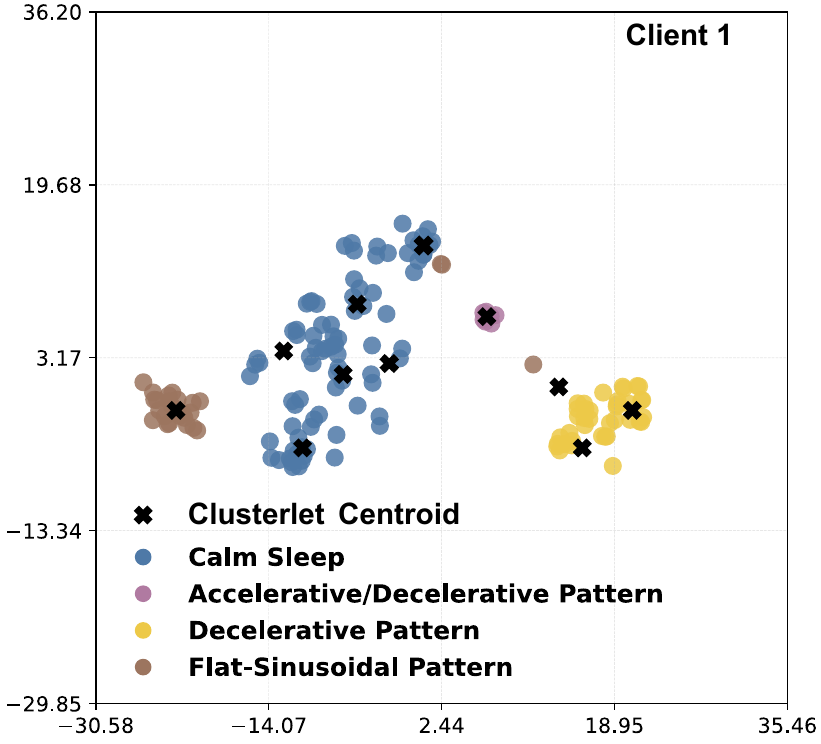}
    \includegraphics[width=0.49\linewidth]{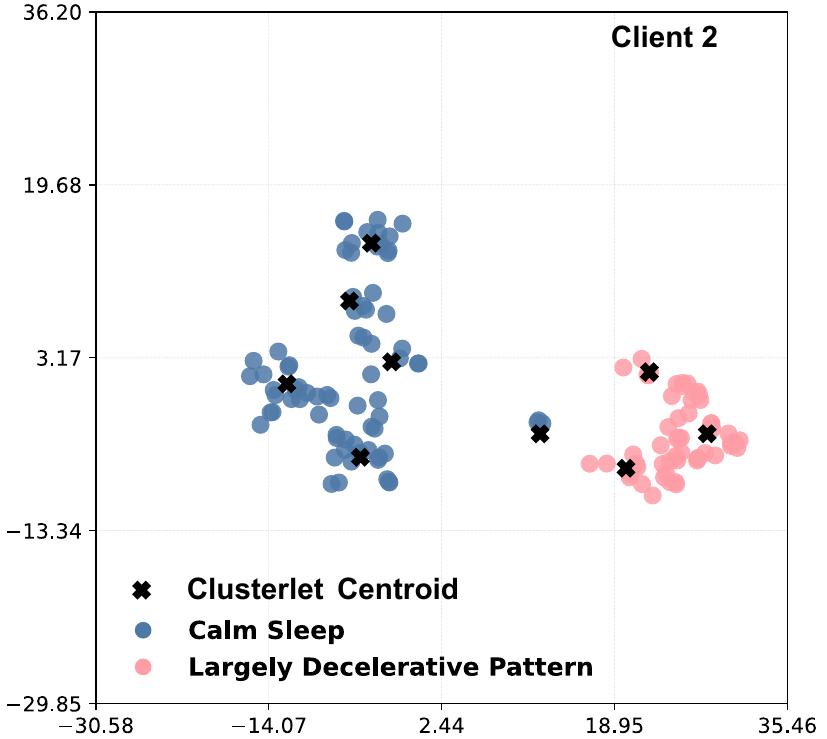}
    \caption{Client-side visualization of Fine-CD on the CA dataset.}
    \label{fig:case_study_client}
\end{figure}

\begin{table}[!t]
    \centering
    \caption{Ablation study of FedHCA components. \textbf{F} and \textbf{M} denote the Fine-CD and Multi-CA components, respectively. Results in \colorbox[RGB]{230, 240, 230}{Light Green} indicate improvement over the baseline, while those in \colorbox[RGB]{200, 215, 200}{Dark Green} denote the best performance.}
    \label{tbl:ablation_component}
    %\resizebox{\columnwidth}{!}{
    \begin{tabular}{c|cc|cccc}
    \toprule
    \textbf{Data} & \textbf{F} & \textbf{M} & \textbf{Purity} & \textbf{ARI} & \textbf{NMI} & \textbf{ACC} \\
    \midrule

    \multirow{4}{*}{EC}
    & \ding{55} & \ding{55} & 0.7248 & 0.5381 & 0.6381 & 0.7125 \\
    & \ding{51} & \ding{55} & \cellcolor[RGB]{230, 240, 230}0.7462 & \cellcolor[RGB]{230, 240, 230}0.6438 & 0.6244 & \cellcolor[RGB]{230, 240, 230}0.7339 \\
    & \ding{55} & \ding{51} & \cellcolor[RGB]{230, 240, 230}0.7615 & \cellcolor[RGB]{230, 240, 230}0.5696 & \cellcolor[RGB]{230, 240, 230}0.6440 & 0.6881 \\
    & \ding{51} & \ding{51} & \cellcolor[RGB]{200, 215, 200}0.7798 & \cellcolor[RGB]{200, 215, 200}0.6881 & \cellcolor[RGB]{200, 215, 200}0.6475 & \cellcolor[RGB]{200, 215, 200}0.7431 \\

    \midrule
    \multirow{4}{*}{EP}
    & \ding{55} & \ding{55} & 0.3038 & 0.0056 & 0.0125 & 0.3038 \\
    & \ding{51} & \ding{55} & \cellcolor[RGB]{230, 240, 230}0.3149 & \cellcolor[RGB]{230, 240, 230}0.0166 & \cellcolor[RGB]{230, 240, 230}0.0176 & \cellcolor[RGB]{230, 240, 230}0.3093 \\
    & \ding{55} & \ding{51} & \cellcolor[RGB]{230, 240, 230}0.3204 & \cellcolor[RGB]{230, 240, 230}0.0091 & \cellcolor[RGB]{230, 240, 230}0.0196 & 0.3022 \\
    & \ding{51} & \ding{51} & \cellcolor[RGB]{200, 215, 200}0.3244 & \cellcolor[RGB]{200, 215, 200}0.0174 & \cellcolor[RGB]{200, 215, 200}0.0248 & \cellcolor[RGB]{200, 215, 200}0.3180 \\

    \midrule
    \multirow{4}{*}{YE}
    & \ding{55} & \ding{55} & 0.4192 & 0.1541 & 0.3159 & 0.3937 \\
    & \ding{51} & \ding{55} & \cellcolor[RGB]{230, 240, 230}0.4658 & \cellcolor[RGB]{230, 240, 230}0.2040 & \cellcolor[RGB]{230, 240, 230}0.3467 & \cellcolor[RGB]{230, 240, 230}0.4541 \\
    & \ding{55} & \ding{51} & \cellcolor[RGB]{230, 240, 230}0.5189 & \cellcolor[RGB]{230, 240, 230}0.2164 & \cellcolor[RGB]{230, 240, 230}0.3374 & \cellcolor[RGB]{230, 240, 230}0.4367 \\
    & \ding{51} & \ding{51} & \cellcolor[RGB]{200, 215, 200}0.5266 & \cellcolor[RGB]{200, 215, 200}0.2449 & \cellcolor[RGB]{200, 215, 200}0.3661 & \cellcolor[RGB]{200, 215, 200}0.4753 \\

    \midrule
    \multirow{4}{*}{CA}
    & \ding{55} & \ding{55} & 0.5064 & 0.1934 & 0.3588 & 0.4047 \\
    & \ding{51} & \ding{55} & \cellcolor[RGB]{230, 240, 230}0.5254 & \cellcolor[RGB]{230, 240, 230}0.2168 & \cellcolor[RGB]{230, 240, 230}0.3831 & \cellcolor[RGB]{230, 240, 230}0.4286 \\
    & \ding{55} & \ding{51} & \cellcolor[RGB]{230, 240, 230}0.5647 & \cellcolor[RGB]{230, 240, 230}0.2313 & \cellcolor[RGB]{230, 240, 230}0.4065 & \cellcolor[RGB]{230, 240, 230}0.4132 \\
    & \ding{51} & \ding{51} & \cellcolor[RGB]{200, 215, 200}0.5996 & \cellcolor[RGB]{200, 215, 200}0.2798 & \cellcolor[RGB]{200, 215, 200}0.4366 & \cellcolor[RGB]{200, 215, 200}0.4629 \\

    \bottomrule
    \end{tabular}
    %}
\end{table}

Fig.~\ref{fig:case_study_client} presents the t-SNE visualization of Fine-CD on two clients from the CA dataset.
Different colored points denote data objects from these biomedical patterns, and black crosses denote the discovered clusterlet centroids.
On Client 1, several patterns coexist, and the Flat-sinusoidal Pattern data are fragmented into clusterlets. 
Accordingly, Fine-CD discovers multiple clusterlet centroids to represent these structures. 
In contrast, Client 2 observes fewer patterns but more compact local structures. 
This observation verifies that local clients may hold incomplete and granularity-misaligned cluster distributions. 
As a result, the learned clusterlets serve as informative prototypes that summarize incomplete client-side distributions and provide useful structural information for subsequent server-side multi-granular aggregation.

\subsection{Ablation Study}
\textbf{Ablation Study of Core Components:}
We assess the effectiveness of the two core components of FedHCA, i.e., Fine-CD (client) and Multi-CA (server). Table~\ref{tbl:ablation_component} reports the results on four biomedical datasets.
In the table, the symbol ``\ding{51}'' indicates that the component is enabled, while ``\ding{55}'' denotes its substitution with a conventional $k$-means algorithm.
For both client- and server-side $k$-means replacements, the number of clusters is set to the ground-truth cluster number $k^*$ reported in Table~\ref{tbl:dataset}.
The experimental results reveal two key findings:
1) Most variants incorporating either Fine-CD or Multi-CA individually outperform the baseline that lacks both components, highlighting the distinct contribution of each module.
2) The variant employing both components consistently achieves the best performance across all validity indices, demonstrating the complementarity and synergistic effect between Fine-CD and Multi-CA.
The results confirm the essential role of these two components in maintaining the effectiveness of FedHCA.

\begin{figure}[!t]
    \centering
    \includegraphics[width=0.49\linewidth]{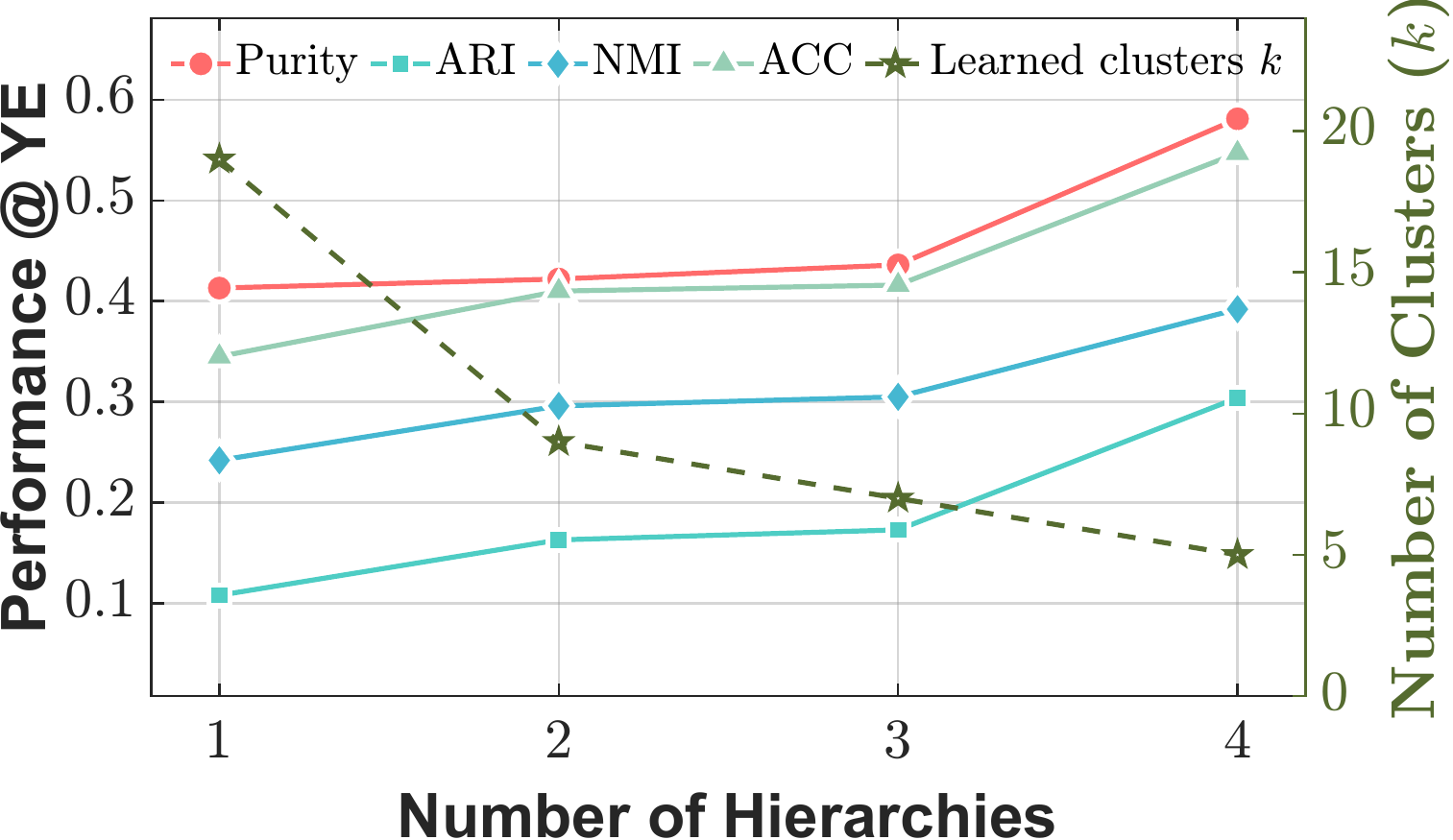}
    \includegraphics[width=0.49\linewidth]{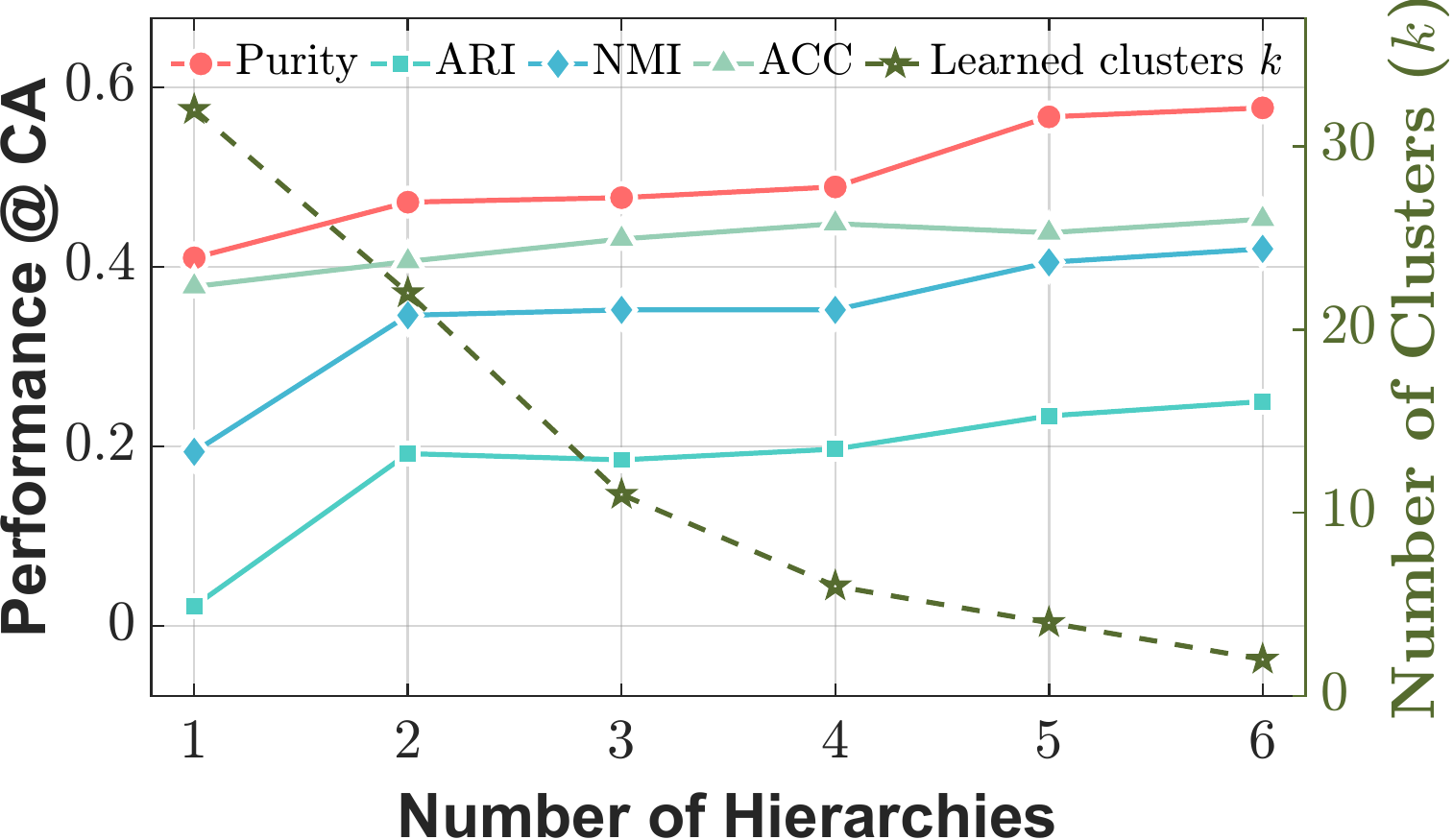}
    \caption{Ablation study of granularity level. As more hierarchies are incorporated, the clustering performance improves.}
    \label{fig:ablation_granularity}
\end{figure}

\begin{figure}[!t]
    \centering
    \includegraphics[width=0.49\linewidth]{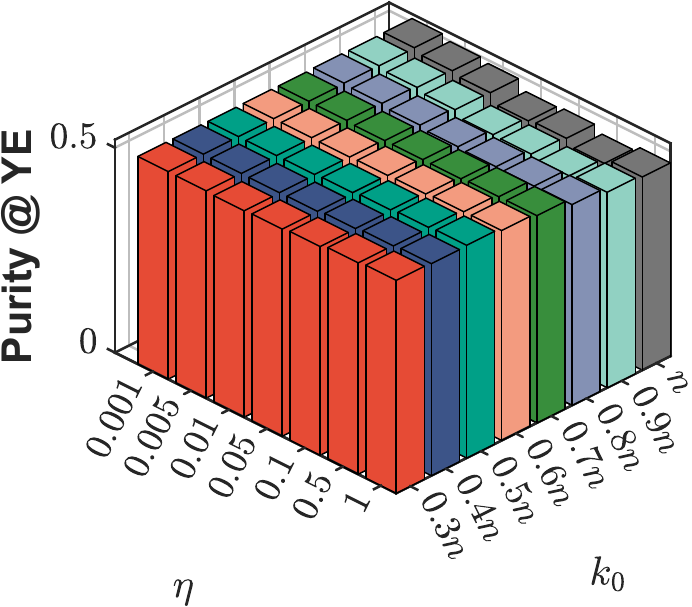}
    \includegraphics[width=0.49\linewidth]{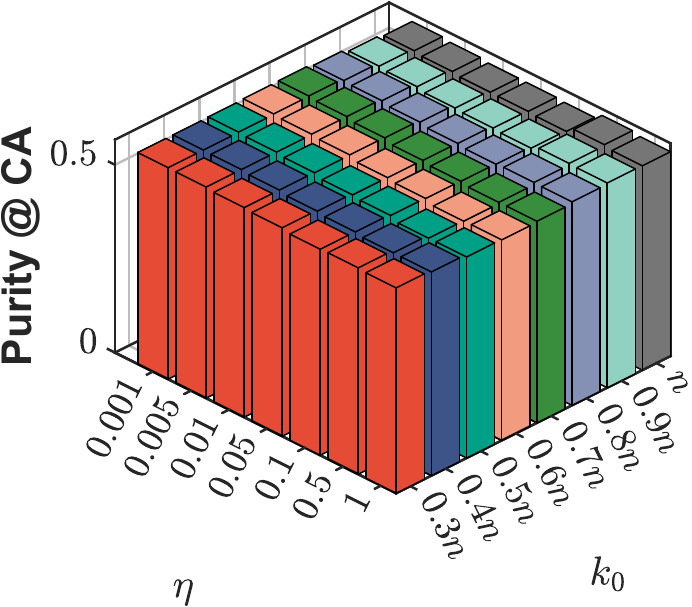}
    \caption{FedHCA performance under $\eta$-$k_0$ combinations.}
    \label{fig:hyper-main}
\end{figure}

\textbf{Ablation Study of Granularity:}
We evaluate the effectiveness of the proposed multi-granular hierarchical structure on two biomedical datasets in Fig.~\ref{fig:ablation_granularity}.
To quantify the contribution of hierarchical granularities, we compare FedHCA with its granularity-restricted variants. 
In the figure, the right axis shows the converged number of clusters at each granularity level, while the left axis reports the performance on four indices.
Specifically, we vary the number of hierarchies involved in the clustering, ranging from only the finest-grained hierarchy to all hierarchical levels.
It can be observed that clustering converges progressively across granularity levels, and the utilization of more levels consistently improves clustering accuracy.
This trend substantiates two critical insights:
1) The granularity diversity provides richer data representations that enhance the clustering.
2) Each granularity level contributes essential information to the clustering task, as evidenced by performance degradation when any level is ablated.

\subsection{Parameter Sensitivity Analysis}
\label{sec:parameter}
% 介绍这个实验
To evaluate hyperparameter sensitivity, we test various combinations of the learning rate $\eta$ and initial cluster count $k_0$. 
%Since Fine-CD and Multi-CA require a large initial size, we exclude $k_0 \in \{0.1n, 0.2n\}$. 
Purity results on two biomedical datasets are shown in Fig.~\ref{fig:hyper-main}. FedHCA demonstrates remarkable robustness to most configurations. Performance only drops under extreme values (e.g., $\eta \in \{0.001, 1\}$ or $k_0 \in \{0.3n, n\}$). It may be due to an overly large $\eta$ or $k_0$, which causes instability and convergence failures. Conversely, a tiny $\eta$ risks trapping the model in local optima, while a small $k_0$ fails to provide sufficient structural information for the clustering task.

\section{Concluding Remarks}

This paper introduces FedHCA to advance FC in realistic biomedical scenarios, where global data topologies are often fragmented across institutions with heterogeneous local distributions. The core innovation lies in synergizing highly efficient one-shot prototype transmission with multi-granular competitive learning. This design elegantly resolves the inherent tension between communication bottlenecks and structural information loss, automatically aggregating incomplete local clusterlets into a globally coherent hierarchy. By further coupling this with an encoding mechanism, FedHCA successfully embeds these multi-granular clustering priors into data-enhanced representations for robust federated cluster analysis. Ultimately, FedHCA effectively reveals comprehensive global data distributions with minimal communication overhead, providing a privacy-preserving and robust tool for distributed biomedical knowledge discovery. Having demonstrated robust performance on static tabular datasets, our future work will focus on adapting this hierarchical FC paradigm to unstructured and complex biomedical data, including non-IID medical imaging and unaligned EEG time series.

\section*{Acknowledgment}
This work was supported in part by the National Natural Science Foundation of China (NSFC) under Grant 62476063.

\bibliographystyle{IEEEtran}
\bibliography{references}

\end{document}